\documentstyle[aclap,psfig]{article}

\title{\vspace{-0.5in}Directed Replacement}
\author{Lauri Karttunen \\
Rank Xerox Research Centre Grenoble \\ 6, chemin de Maupertuis \\
{\sc f-{\small 38240} meylan, france} \\
{\tt lauri.karttunen@xerox.fr}}

\begin{document}
\bibliographystyle{fullname}
\maketitle
\vspace{-0.5in}
\begin{abstract}

This paper introduces to the finite-state calculus a family of
directed replace operators. In contrast to the simple replace
expression, {\tt UPPER -> LOWER}, defined in Karttunen (1995), the new
directed version, {\tt UPPER @-> LOWER}, yields an unambiguous
transducer if the lower language consists of a single string. It
transduces the input string from left to right, making only the
longest possible replacement at each point.

A new type of replacement expression, {\tt UPPER @-> PREFIX
... SUFFIX}, yields a transducer that inserts text around strings that
are instances of {\tt UPPER}.  The symbol ... denotes the matching
part of the input which itself remains unchanged. {\tt PREFIX} and
{\tt SUFFIX} are regular expressions describing the insertions.

Expressions of the type {\tt UPPER @-> PREFIX ... SUFFIX} may be used
to compose a deterministic parser for a ``local grammar'' in the sense
of Gross (1989). Other useful applications of directed replacement
include tokenization and filtering of text streams.
\end{abstract}

\section{Introduction}
\begin{figure}
\begin{center}
  \centerline{\psfig{file=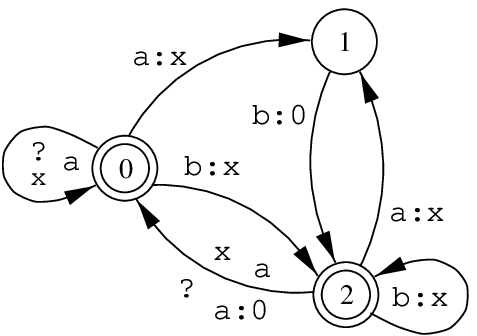}}
\caption{\label{net1}\verb+ a b | b | b a | a b a -> x +. The four
paths with ``aba'' on the upper side are: $<$0~{\tt a}~0~{\tt
b:x}~2~{\tt a}~0$>$, $<$0~{\tt a}~0~{\tt b:x}~2~{\tt a:0}~0$>$,
$<$0~{\tt a:x}~1~{\tt b:0}~2~{\tt a}~0$>$, and $<$0~{\tt a:x}~1~{\tt
b:0}~2~{\tt a:0}~0$>$.}
\end{center}
\vspace*{-8mm}
\end{figure}

Transducers compiled from simple replace expressions {\tt UPPER} {\tt
->} {\tt LOWER} (Karttunen 1995, Kempe and Karttunen 1996) are
generally nondeterministic in the sense that they may yield multiple
results even if the lower language consists of a single string.  For
example, let us consider the transducer in Figure \ref{net1},
representing {\tt a b | b | b a | a b a -> x}.\footnote{The regular
expression formalism and other notational conventions used in
the paper are explained in the Appendix at the end.}

The application of this transducer to the input ``aba'' produces
four alternate results, ``axa'', ``ax'', ``xa'', and ``x'', as shown
in Figure \ref{net1}, since there are four paths in the network that
contain ``aba'' on the upper side with different strings on the lower
side.

\begin{figure}[here]
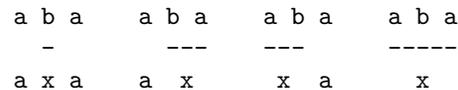

\vspace*{-2mm}
\begin{verbatim}
    a b a    a b a    a b a    a b a
      -        ---    ---      -----
    a x a    a  x      x  a      x
\end{verbatim}

\caption{\label{tab1}Four factorizations of ``aba''.}
\vspace*{-2mm}
\end{figure}

This nondeterminism arises in two ways. First of all, a replacement
can start at any point.  Thus we get different results for ``aba''
depending on whether we start at the beginning of the string or in the
middle at the ``b''. Secondly, there may be alternative replacements
with the same starting point. In the beginning of ``aba'', we can
replace either ``ab'' or ``aba''. Starting in the middle, we can
replace either ``b'' or ``ba''.  The underlining in Figure \ref{tab1}
shows the four alternate factorizations of the input string, that is,
the four alternate ways to partition the string ``aba'' with respect
to the upper language of the replacement expression. The corresponding
paths in the transducer are listed in Figure \ref{net1}.

For many applications, it is useful to define another version of
replacement that produces a unique outcome whenever the lower language
of the relation consists of a single string. To limit the number of
alternative results to one in such cases, we must impose a unique
factorization on every input.

The desired effect can be obtained by constraining the directionality
and the length of the replacement.  Directionality means that the
replacement sites in the input string are selected starting from the
left or from the right, not allowing any overlaps. The length
constraint forces us always to choose the longest or the shortest
replacement whenever there are multiple candidate strings starting at
a given location. We use the term {\bf directed replacement} to
describe a replacement relation that is constrained by directionality
and length of match. (See the end of Section 2 for a discussion about
the choice of the term.)

With these two kinds of constraints we can define four types of
directed replacement, listed in Figure \ref{tab2}.

\begin{figure}[here]
\vspace*{-2mm}
\begin{verbatim}
                    longest     shortest
                     match        match
  left-to-right       @->          @>
  right-to-left       ->@          >@
\end{verbatim}
\caption{\label{tab2}Directed replacement operators}
\vspace*{-2mm}
\end{figure}

For reasons of space, we discuss here only the left-to-right,
longest-match version. The other cases are similar.

\begin{figure}[here]
\begin{center}
  \centerline{\psfig{file=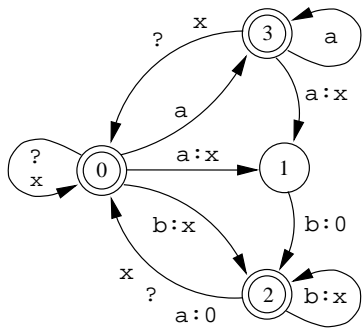}}
\caption{\label{net2}\verb+a b | b | b a | a b a @-> x+. The single
path with ``aba'' on the upper side is: $<$0~{\tt a:x}~1~{\tt b:0}~2
{\tt a:0}~0$>$.}
\end{center}
\vspace*{-6mm}
\end{figure}

The effect of the directionality and length constraints is that some
possible replacements are ignored. For example, {\tt  a b | b | b a |
a b a @-> x } maps ``aba'' uniquely into ``x'', Figure \ref{net2}.

Because we must start from the left and have to choose the longest
match, ``aba'' must be replaced, ignoring the possible replacements
for ``b'', ``ba'', and ``ab''. The {\tt @->} operator allows only the
last factorization of ``aba'' in Figure \ref{tab1}.

\begin{figure}[here]
\vspace*{-2mm}
\begin{center}
  \centerline{\psfig{file=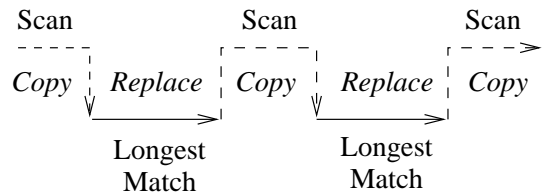}}
\caption{\label{pict1}Left-to-right, longest-match replacement}
\end{center}
\vspace*{-6mm}
\end{figure}

Left-to-right, longest-match replacement can be thought of as a
procedure that rewrites an input string sequentially from left to
right. It copies the input until it finds an instance of {\tt
UPPER}. At that point it selects the longest matching substring, which
is rewritten as {\tt LOWER}, and proceeds from the end of that
substring without considering any other alternatives. Figure
\ref{pict1} illustrates the idea.

It is not obvious at the outset that the operation can in fact be
encoded as a finite-state transducer for arbitrary regular patterns.
Although a unique substring is selected for replacement at each point,
in general the transduction is not unambiguous because {\tt LOWER} is
not required to be a single string; it can be any regular language.

The idea of treating phonological rewrite rules in this way was the
starting point of Kaplan and Kay (1994). Their notion of obligatory
rewrite rule incorporates a directionality constraint. They observe
(p. 358), however, that this constraint does not by itself guarantee a
single output. Kaplan and Kay suggest that additional restrictions,
such as longest-match, could be imposed to further constrain rule
application.\footnote{\label{foot1}The tentative formulation of the
longest-match constraint in \cite[p. 358]{Kaplan+Kay:regmod} is too
weak. It does not cover all the cases.} We consider this issue in more
detail.

The crucial observation is that the two constraints, left-to-right and
longest-match, force a unique factorization on the input string thus
making the transduction unambiguous if the {\tt LOWER} language
consists of a single string. In effect, the input string is
unambiguously {\bf parsed} with respect to the {\tt UPPER} language.
This property turns out to be important for a number of applications.
Thus it is useful to provide a replacement operator that implements
these constraints directly.

The definition of the \verb.UPPER @-> LOWER. relation is presented in
the next section.  Section 3 introduces a novel type of replace
expression for constructing transducers that unambiguously recognize
and mark instances of a regular language without actually replacing
them.  Section 4 identifies some useful applications of the new
replacement expressions.

\section{Directed Replacement}
We define directed replacement by means of a composition of regular
relations. As in Kaplan and Kay (1994), Karttunen (1995), and other
previous works on related topics, the intermediate levels of the
composition introduce auxiliary symbols to express and enforce
constraints on the replacement relation. Figure \ref{tab3} shows the
component relations and how they are composed with the input.

\begin{figure}[here]
\vspace*{-2mm}
\begin{verbatim}
          Input string
              .o.
          Initial match
              .o.
          Left-to-right constraint
              .o.
          Longest-match constraint
              .o.
          Replacement
\end{verbatim}
\caption{\label{tab3}Composition of directed replacement}
\vspace*{-2mm}
\end{figure}

If the four relations on the bottom of Figure \ref{tab3} are composed in
advance, as our compiler does, the application of the replacement to
an input string takes place in one step without any intervening levels
and with no auxiliary symbols. But it helps to understand the logic to
see where the auxiliary marks would be in the hypothetical
intermediate results.

Let us consider the case of {\tt  a b | b | b a | a b a} {\tt @-> x }
applying to the string ``aba'' and see in detail how the mapping
implemented by the transducer in Figure \ref{net2} is composed from
the four component relations. We use three auxiliary symbols, caret
(\verb+^+), left bracket (\verb+<+) and right bracket (\verb+>+),
assuming here that they do not occur in any input. The first step,
shown in Figure \ref{tab4}, composes the input string with a
transducer that inserts a caret, in the beginning of every substring
that belongs to the upper language.

\begin{figure}[here]
\vspace*{-2mm}
\begin{verbatim}
            a    b  a  
          ^ a ^  b  a
\end{verbatim}
\caption{\label{tab4}Initial match. Each caret marks the beginning of
a substring that matches ``ab'', ``b'', ``ba'', or ``aba''.}
\vspace*{-2mm}
\end{figure}

Note that only one \verb+^+ is inserted even if there are several
candidate strings starting at the same location.

In the left-to-right step, we enclose in angle brackets all the
substrings starting at a location marked by a caret that are instances
of the upper language. The initial caret is replaced by a {\tt <}, and
a closing {\tt >} is inserted to mark the end of the match. We permit
carets to appear freely while matching. No carets are permitted
outside the matched substrings and the ignored internal carets are
eliminated. In this case, there are four possible outcomes, shown in
Figure \ref{tab5}, but only two of them are allowed under the
constraint that there can be no carets outside the brackets.

\begin{figure}[here]
\vspace*{-2mm}
\begin{verbatim}
              ALLOWED

    ^ a ^ b   a     ^ a ^ b   a
    < a   b > a     < a   b   a >

            NOT ALLOWED

    ^ a  ^ b   a    ^ a ^ b   a
    ^ a  < b > a    ^ a < b   a >  
\end{verbatim}
\caption{\label{tab5}Left-to-right constraint. {\it No caret
outside a bracketed region.}}
\vspace*{-2mm}
\end{figure}

In effect, no starting location for a replacement can be skipped over
except in the context of another replacement starting further left in
the input string. (Roche and Schabes (1995) introduce a similar
technique for imposing the left-to-right order on the transduction.)
Note that the four alternatives in Figure \ref{tab5} represent the four
factorizations in Figure \ref{tab1}.

The longest-match constraint is the identity relation on a certain set
of strings. It forbids any replacement that starts at the same
location as another, longer replacement. In the case at hand, it means
that the internal {\tt >} is disallowed in the context \verb+< a b >
a+.  Because ``aba'' is in the upper language, there is a longer, and
therefore preferred, \verb+< a b a >+ alternative at the same starting
location, Figure \ref{tab5a}.

\begin{figure}[here]
\vspace*{-2mm}
\begin{verbatim}
       ALLOWED         NOT ALLOWED

    < a   b   a >      < a   b > a     
\end{verbatim}
\caption{\label{tab5a}Longest match constraint. {\it No upper language
string with an initial }{\tt < }{\it and a nonfinal }{\tt > }{\it in
the middle}.}
\vspace*{-2mm}
\end{figure}

In the final replacement step, the bracketed regions of the input
string, in the case at hand, just \verb+< a b a >+ , are replaced by
the strings of the lower language, yielding ``x'' as the result for
our example.

Note that longest match constraint ignores any internal brackets. For
example, the bracketing {\tt < a > < a >} is not allowed if the upper
language contains ``aa'' as well as ``a''. Similarly, the
left-to-right constraint ignores any internal carets.

As the first step towards a formal definition of {\tt UPPER @-> LOWER}
it is useful to make the notion of ``ignoring internal brackets'' more
precise. Figure \ref{tab5b} contains the auxiliary definitions. For
the details of the formalism (briefly explained in the Appendix),
please consult Karttunen (1995), Kempe and Karttunen
(1996).\footnote{\label{foot2}{\tt UPPER'} is the same language as
{\tt UPPER} except that carets may appear freely in all nonfinal
positions. Similarly, {\tt UPPER''} accepts any nonfinal brackets.}

\begin{figure}[here]
\vspace*{-2mm}
\begin{verbatim}
  UPPER'  = UPPER/[%^] - [?* %^]
  UPPER'' = UPPER/[%<|%>] - [?* [%<|%>]]
\end{verbatim}
\caption{\label{tab5b}Versions of {\tt UPPER} that freely allow
nonfinal diacritics.}
\vspace*{-2mm}
\end{figure}    

The precise definition of the \verb+UPPER @-> LOWER+ relation is given
in Figure \ref{tab6}. It is a composition of many auxiliary relations.
We label the major components in accordance with the outline in Figure
\ref{tab3}. The formulation of the longest-match constraint is based
on a suggestion by Ronald M. Kaplan (p.c.).
\begin{figure}[here]
\vspace*{-2mm}
{\it Initial match}
\begin{verbatim}
        ~$[ %^ | %< | %> ]
               .o.
      [. .] -> %^ || _ UPPER
               .o.
\end{verbatim}
{\it Left to right}
\begin{verbatim}
  [~$[%^] [%^:%< UPPER' 0:%>]]* ~$[%^]
               .o.
             %^ -> []
               .o.
\end{verbatim}
{\it Longest match}
\begin{verbatim}
      ~$[%< [UPPER'' & $[%>]]]
               .o.
\end{verbatim}
{\it Replacement}
\begin{verbatim}
        %< ~$[%>] %> -> LOWER  ;
\end{verbatim}
\caption{\label{tab6}Definition of {\tt UPPER @-> LOWER}}
\vspace*{-2mm}
\end{figure}

The logic of {\tt @->} replacement could be encoded in many other
ways, for example, by using the three pairs of auxiliary brackets,
{\tt <i}, {\tt >i}, {\tt <c}, {\tt >c}, and {\tt <a}, {\tt >a},
introduced in Kaplan and Kay (1994). We take here a more minimalist
approach. One reason is that we prefer to think of the simple
unconditional (uncontexted) replacement as the basic case, as in
Karttunen (1995). Without the additional complexities introduced by
contexts, the directionality and length-of-match constraints can be
encoded with fewer diacritics. (We believe that the conditional case
can also be handled in a simpler way than in Kaplan and Kay (1994).)
The number of auxiliary markers is an important consideration for some
of the applications discussed below.

In a phonological or morphological rewrite rule, the center part of
the rule is typically very small: a modification, deletion or
insertion of a single segment. On the other hand, in our text
processing applications, the upper language may involve a large
network representing, for example, a lexicon of multiword
tokens. Practical experience shows that the presence of many auxiliary
diacritics makes it difficult or impossible to compute the
left-to-right and longest-match constraints in such cases. The size of
intermediate states of the computation becomes a critical issue, while
it is irrelevant for simple phonological rules.  We will return to
this issue in the discussion of tokenizing transducers in Section 4.

The transducers derived from the definition in Figure \ref{tab6} have
the property that they unambiguously parse the input string into a
sequence of substrings that are either copied to the output unchanged or
replaced by some other strings. However they do not fall neatly into
any standard class of transducers discussed in the literature
(Eilenberg 1974, Sch\"{u}tzenberger 1977, Berstel 1979). If the {\tt
LOWER} language consists of a single string, then the relation encoded
by the transducer is in Berstel's terms a {\bf rational function}, and
the network is an {\bf unambigous} transducer, even though it may
contain states with outgoing transitions to two or more destinations
for the same input symbol. An unambiguous transducer may also be {\bf
sequentiable}, in which case it can be turned into an equivalent {\bf
sequential} transducer \cite{Mohri:fsa+nlp}, which can in turn be
minimized. A transducer is sequential just in case there are no states
with more than one transition for the same input symbol. Roche and
Schabes (1995) call such transducers {\bf deterministic}.

\begin{figure}
\begin{center}
  \centerline{\psfig{file=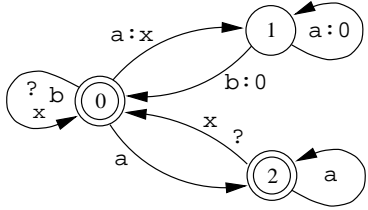}}
\caption{\label{net3}\verb| a+ b @-> x|. This transducer is
unambiguous but cannot be sequentialized.}
\end{center}
\vspace*{-8mm}
\end{figure}

Our replacement transducers in general are not unambiguous because we
allow {\tt LOWER} to be any regular language. It may well turn out
that, in all cases that are of practical interest, the lower language
is in fact a singleton, or at least some finite set, but it is not so
by definition. Even if the replacement transducer is unambiguous, it
may well be unsequentiable if {\tt UPPER} is an infinite language.
For example, the simple transducer for {\tt a+ b @-> x} in Figure
\ref{net3} cannot be sequentialized. It has to replace any string of
``a''s by ``x'' or copy it to the output unchanged depending on
whether the string eventually terminates at ``b''. It is obviously
impossible for any finite-state device to accumulate an unbounded
amount of delayed output. On the other hand, the transducer in Figure
\ref{net2} is sequentiable because there the choice between {\tt a}
and {\tt a:x} just depends on the next input symbol.

Because none of the classical terms fits exactly, we have chosen a
novel term, {\bf directed transduction}, to describe a relation
induced by the definition in Figure \ref{tab6}. It is meant to suggest
that the mapping from the input into the output strings is guided by the
directionality and length-of-match constraints. Depending on the
characteristics of the {\tt UPPER} and {\tt LOWER} languages, the
resulting transducers may be unambiguous and even sequential, but
that is not guaranteed in the general case.

\section{Insertion}
The effect of the left-to-right and longest-match constraint is to
factor any input string uniquely with respect to the upper language of
the replace expression, to parse it into a sequence of substrings that
either belong or do not belong to the language. Instead of replacing
the instances of the upper language in the input by other strings, we
can also take advantage of the unique factorization in other ways. For
example, we may insert a string before and after each substring that
is an instance of the language in question simply to mark it as such.

To implement this idea, we introduce the special symbol ... on the
right-hand side of the replacement expression to mark the place around
which the insertions are to be made. Thus we allow replacement
expressions of the form {\tt UPPER @-> PREFIX ... SUFFIX}. The
corresponding transducer locates the instances of {\tt UPPER} in the
input string under the left-to-right, longest-match regimen just
described. But instead of replacing the matched strings, the
transducer just copies them, inserting the specified prefix and
suffix.  For the sake of generality, we allow {\tt PREFIX} and {\tt
SUFFIX} to denote any regular language.

The definition of {\tt UPPER @-> PREFIX ...} {\tt SUFFIX} is just as
in Figure \ref{tab6} except that the Replacement expression is
replaced by the Insertion formula in Figure \ref{tab7}, a simple
parallel replacement of the two auxiliary brackets that mark the
selected regions. Because the placement of \verb+<+ and \verb+>+ is
strictly controlled, they do not occur anywhere else.

\begin{figure}[here]
\vspace*{-2mm}
{\it Insertion}
\begin{verbatim}
        %< -> PREFIX, %> -> SUFFIX ;
\end{verbatim}
\caption{\label{tab7}Insertion expression in the definition of
{\tt UPPER @-> PREFIX ... SUFFIX}.}
\vspace*{-2mm}
\end{figure}

With the ... expressions we can construct transducers that mark
maximal instances of a regular language. For example, let us assume
that noun phrases consist of an optional determiner, {\tt (d)}, any
number of adjectives, {\tt a*}, and one or more nouns, {\tt n+}. The
expression \verb| (d) a* n+ @-> %[ ... %] | compiles into a transducer
that inserts brackets around maximal instances of the noun phrase
pattern.  For example, it maps {\tt "dannvaan"} into {\tt
"[dann]v[aan]"}, as shown in Figure \ref{tab8}.

\begin{figure}[here]
\vspace*{-2mm}
\begin{verbatim}
        d a n n   v   a a n
        -------       -----
      [ d a n n ] v [ a a n ]
\end{verbatim}
\caption{\label{tab8}Application of\verb| (d) a* n+ @-> \%[...\%] |to
{\tt "dannvaan"}}
\vspace*{-2mm}
\end{figure}

\begin{figure}[here]
\vspace*{-2mm}
\begin{center}
  \centerline{\psfig{file=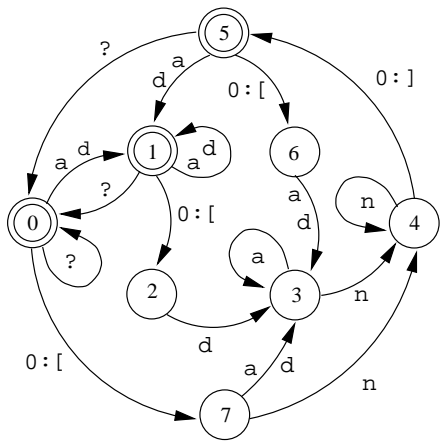}}
\caption{\label{net4}\verb|(d) a* n+ @-> \%[...\%]|. The
one path with ``dannvaan'' on the upper side is: $<$0 {\tt 0:[} 7
{\tt d} 3 {\tt a} 3 {\tt n} 4 {\tt n} 4 {\tt 0:]} 5 {\tt v} 0 {\tt
0:[} 7 {\tt a} 3 {\tt a} 3 {\tt n} 4 {\tt 0:]} 5$>$.}
\end{center}
\vspace*{-6mm}
\end{figure}

Although the input string \verb+"dannvaan"+ contains many other
instances of the noun phrase pattern, \verb+"n"+, \verb+"an"+,
\verb+"nn"+, etc., the left-to-right and longest-match constraints
pick out just the two maximal ones. The transducer is displayed
in Figure \ref{net4}. Note that {\tt ?} here matches symbols, such as
{\tt v}, that are not included in the alphabet of the network.

\section{Applications}
The directed replacement operators have many useful applications.  We
describe some of them. Although the same results could often be
achieved by using lex and yacc, sed, awk, perl, and other Unix
utilities, there is an advantage in using finite-state transducers for
these tasks because they can then be smoothly integrated with other
finite-state processes, such as morphological analysis by lexical
transducers (Karttunen 1994) and rule-based part-of-speech
disambiguation (Chanod and Tapanainen 1995, Roche and Schabes 1995).

\subsection{Tokenization}
A tokenizer is a device that segments an input string into a sequence
of tokens. The insertion of end-of-token marks can be accomplished by
a finite-state transducer that is compiled from tokenization
rules. The tokenization rules may be of several types. For example,
\verb|[WHITE%_SPACE+| \verb|@-> SPACE]| is a normalizing transducer
that reduces any sequence of tabs, spaces, and newlines to a single
space. \verb|[LETTER+| \verb|@-> ...  END%_OF%_TOKEN]| inserts a
special mark, e.g. a newline, at the end of a letter sequence.

Although a space generally counts as a token boundary, it can also be
part of a multiword token, as in expressions like ``at least'', ``head
over heels'', ``in spite of'', etc. Thus the rule that introduces the
\verb+END_OF_TOKEN+ symbol needs to combine the \verb|LETTER+| pattern
with a list of multiword tokens which may include spaces, periods and
other delimiters.

Figure \ref{tab9} outlines the construction of a simple tokenizing transducer
for English.

\begin{figure}[here]
\vspace{-2mm}
\begin{verbatim}
   WHITE%_SPACE+ @-> SPACE
            .o.
   [ LETTER+ | 
     a t %  l e a s t |
     h e a d %  o v e r %  h e e l s |
     i n %  s p i t e %  o f ]
    @-> ... END%_OF%_TOKEN
            .o.
   SPACE -> [] || .#. | END%_OF%_TOKEN _ ;
\end{verbatim}
\caption{\label{tab9}A simple tokenizer}
\vspace{-2mm}
\end{figure}

The tokenizer in Figure \ref{tab9} is composed of three
transducers. The first reduces strings of whitespace characters to a
single space. The second transducer inserts an \verb+END_OF_TOKEN+
mark after simple words and the listed multiword expressions. The
third removes the spaces that are not part of some multiword
token. The percent sign here means that the following blank is to be
taken literally, that is, parsed as a symbol.

Without the left-to-right, longest-match constraints, the tokenizing
transducer would not produce deterministic output.  Note that it must
introduce an \verb+END_OF_TOKEN+ mark after a sequence of letters just
in case the word is not part of some longer multiword token. This
problem is complicated by the fact that the list of multiword tokens
may contain overlapping expressions. A tokenizer for French, for
example, needs to recognize ``de plus'' (moreover), ``en plus''
(more), ``en plus de'' (in addition to), and ``de plus en plus'' (more
and more) as single tokens.  Thus there is a token boundary after ``de
plus'' in {\it de plus on ne le fait plus} (moreover one doesn't do it
anymore) but not in {\it on le fait de plus en plus} (one does it more
and more) where ``de plus en plus'' is a single token.

If the list of multiword tokens contains hundreds of expressions, it
may require a lot of time and space to compile the tokenizer even if
the final result is not too large.  The number of auxiliary symbols
used to encode the constraints has a critical effect on the efficiency
of that computation. We first observed this phenomenon in the course
of building a tokenizer for the British National Corpus according to
the specifications of the {\sc bnc} Users Guide \cite{Leech:bnc},
which lists around 300 multiword tokens and 260 foreign phrases.  With
the current definition of the directed replacement we have now been
able to compute similar tokenizers for several other languages
(French, Spanish, Italian, Portuguese, Dutch, German).

\subsection{Filtering}
Some text processing applications involve a preliminary stage in which
the input stream is divided into regions that are passed on to the
calling process and regions that are ignored. For example, in
processing an {\sc sgml}-coded document, we may wish to delete all the
material that appears or does not appear in a region bounded by
certain {\sc sgml} tags, say {\tt <A>} and {\tt </A>}.

Both types of filters can easily be constructed using the
directed replace operator. A negative filter that deletes all the
material between the two {\sc sgml} codes, including the codes themselves,
is expressed as in Figure \ref{tab10}.

\begin{figure}[here]
\vspace{-2mm}
\begin{verbatim}
   "<A>" ~$["<A>"|"</A>"] "</A>" @-> [] ;
\end{verbatim}
\caption{\label{tab10} A negative filter}
\vspace{-2mm}
\end{figure}

A positive filter that excludes everything else can be expressed as in
Figure \ref{tab11}.

\begin{figure}[here]
\vspace{-2mm}
\begin{verbatim}
         ~$"</A>" "<A>" @-> "<A>"
                 .o.
         "</A>" ~$"<A>" @-> "</A>" ;
\end{verbatim}
\caption{\label{tab11}A positive filter}
\vspace{-2mm}
\end{figure}

The positive filter is composed of two transducers. The first reduces
to {\tt <A>} any string that ends with it and does not contain the
{\tt </A>} tag. The second transducer does a similar transduction on
strings that begin with {\tt </A>}. Figure 12 illustrates the effect
of the positive filter.

\begin{figure}[here]
\vspace{-2mm}
\begin{verbatim}
<B>one</B><A>two</A><C>three</C><A>four</A>
-------------   ----------------
     <A>     two      </A>      <A>four</A>
\end{verbatim}
\caption{\label{tab12}Application of a positive filter}
\vspace{-2mm}
\end{figure}

The idea of filtering by finite-state transduction of course does not
depend on {\sc sgml} codes. It can be applied to texts where the
interesting and uninteresting regions are defined by any kind of
regular pattern.

\subsection{Marking}

As we observed in section 3, by using the ... symbol on the lower side
of the replacement expression, we can construct transducers that mark
instances of a regular language without changing the text in any other
way. Such transducers have a wide range of applications. They can be
used to locate all kinds of expressions that can be described by a
regular pattern, such as proper names, dates, addresses, social
security and phone numbers, and the like. Such a marking transducer
can be viewed as a deterministic parser for a ``local grammar'' in the
sense of Gross (1989), Roche (1993), Silberztein (1993) and others.
 
By composing two or more marking transducers, we can also construct a
single transducer that builds nested syntactic structures, up to any
desired depth. To make the construction simpler, we can start by
defining auxiliary symbols for the basic regular patterns. For example,
we may define {\tt NP} as \verb|[(d) a* n+]|. With that abbreviatory
convention, a composition of a simple {\tt NP} and {\tt VP} spotter
can be defined as in Figure \ref{tab13}.

\begin{figure}[here]
\vspace{-2mm}
\begin{verbatim}
       NP @-> %[NP ... %]
             .o.
       v %[NP NP %]  @-> %[VP ... %] ;
\end{verbatim}
\caption{\label{tab13}Composition of an {\tt NP} and a {\tt VP} spotter}
\vspace{-2mm}
\end{figure}

Figure \ref{tab14} shows the effect of applying this composite transducer to
the string {\tt "dannvaan"}.

\begin{figure}[here]
\vspace{-2mm}
\begin{verbatim}
         d a n n       v     a a n
         -------       -     -----
     [NP d a n n ] [VP v [NP a a n ] ]
\end{verbatim}
\caption{\label{tab14}Application of an {\tt NP-VP} parser}
\vspace{-2mm}
\end{figure}

By means of this simple ``bottom-up'' technique, it is possible to
compile finite-state transducers that approximate a context-free
parser up to a chosen depth of embedding. Of course, the
left-to-right, longest-match regimen implies that some possible
analyses are ignored. To produce all possible parses, we may introduce
the ... notation to the simple replace expressions in Karttunen
(1995).

\section{Extensions}

The definition of the left-to-right, longest-match replacement can
easily be modified for the three other directed replace operators
mentioned in Figure \ref{tab2}. Another extension, already
implemented, is a directed version of parallel replacement (Kempe and
Karttunen 1996), which allows any number of replacements to be done
simultaneously without interfering with each other. Figure \ref{tab15}
is an example of a directed parallel replacement. It yields a
transducer that maps a string of ``a''s into a single ``b'' and 
a string of ``b''s into a single ``a''.

\begin{figure}[here]
\begin{verbatim}
        a+ @-> b, b+ @-> a ;
\end{verbatim}
\caption{\label{tab15}Directed, parallel replacement}
\vspace{-2mm}
\end{figure}

The definition of directed parallel replacement requires no
additions to the techniques already presented. In the near future we
also plan to allow directional and length-of-match constraints in the
more complicated case of conditional context-constrained replacement.

\section{Acknowledgements}

I would like to thank Ronald M. Kaplan, Martin Kay, Andr\'{e} Kempe,
John Maxwell, and Annie Zaenen for helpful discussions at the
beginning of the project, as well as Paula Newman and Kenneth
R. Beesley for editorial advice on the first draft of the paper. The
work on tokenizers and phrasal analyzers by Anne Schiller and Gregory
Grefenstette revealed the need for a more efficient implementation of
the idea. The final version of the paper has benefited from detailed
comments by Ronald M. Kaplan and two anonymous reviewers, who
convinced me to discard the ill-chosen original title (``Deterministic
Replacement'') in favor of the present one.

\section{Appendix: Notational conventions}

The regular expression formalism used in this paper is essentially the
same as in Kaplan and Kay (1994), in Karttunen (1995), and in Kempe
and Karttunen (1996). Upper-case strings, such as {\tt UPPER},
represent regular languages, and lower-case letters, such as {\tt x},
represent symbols. We recognize two types of symbols: unary symbols
({\tt a}, {\tt b}, {\tt c}, etc) and symbol pairs ({\tt a:x}, {\tt
b:0}, etc. ).

A symbol pair {\tt a:x} may be thought of as the crossproduct of {\tt
a} and {\tt x}, the minimal relation consisting of {\tt a} (the upper
symbol) and {\tt x} (the lower symbol). To make the notation less
cumbersome, we systematically ignore the distinction between the
language {\tt A} and the identity relation that maps every string
of {\tt A} into itself. Consequently, we also write {\tt a:a} as
just {\tt a}.

Three special symbols are used in regular expressions: {\tt 0} (zero)
represents the empty string (often denoted by $\epsilon$); {\tt ?}
stands for any symbol in the known alphabet and its extensions; in
replacement expressions, {\tt .\#.} marks the start (left context) or
the end (right context) of a string.  The percent sign, {\tt \%}, is
used as an escape character. It allows letters that have a special
meaning in the calculus to be used as ordinary symbols. Thus {\tt \%[}
denotes the literal square bracket as opposed to {\tt [}, which has a
special meaning as a grouping symbol; \%0 is the ordinary zero symbol.
Double quotes around a symbol have the same effect as the percent
sign.

The following simple expressions appear freqently in the formulas:
{\tt []} the empty string language, {\tt ?*} the universal (``sigma
star'') language.

The regular expression operators used in the paper are: {\tt *} zero
or more (Kleene star), {\tt +} one or more (Kleene plus), \verb+~+ not
(complement), {\tt \$} contains, {\tt /} ignore, {\tt |} or (union),
{\tt \&} and (intersection), {\tt -} minus (relative complement), {\tt
.x.} crossproduct, {\tt .o.} composition, \verb+->+ simple replace.

In the transducer diagrams (Figures \ref{net1}, \ref{net2}, etc.),
the nonfinal states are represented by single circles, final states
by double circles. State 0 is the initial state. The symbol {\tt ?}
represents any symbols that are not explicitly present in the
network. Transitions that differ only with respect to the label
are collapsed into a single multiply labelled arc.


\begin{thebibliography}{}

\end{thebibliography}


\begin{thebibliography}{fullname}

\bibitem[\protect\citename{Berstel}1979]{Berstel:t&cfl}
Jean Berstel.
\newblock 1979.
\newblock {\em Transductions and Context-Free Languages}.
\newblock B.G. Teubner, Stuttgart, Germany.

\bibitem[\protect\citename{Chanod and
Tapanainen}1995]{Chanod+Tapanainen:dub} Jean-Pierre Chanod and Pasi
Tapanainen.  \newblock 1995.  \newblock Tagging French---comparing a
statistical and a constraint-based model.  \newblock In {\em The
Proceedings of the Seventh Conference of the European Chapter of the
Association for Computational Linguistics}, Dublin, Ireland.

\bibitem[\protect\citename{Eilenberg}1974]{Eilenberg:alm}
Samuel Eilenberg.
\newblock 1974.
\newblock {\em Automata, Languages, and Machines}.
\newblock Academic Press.

\bibitem[\protect\citename{Gross}1989]{Gross:fsa}
Maurice Gross.
\newblock 1989.
\newblock The Use of Finite Automata in the Lexical Representation
of Natural Language.
\newblock In {\em Lecture Notes in Computer Science}, pages
  34--50, Springer-Verlag, Berlin, Germany.

\bibitem[\protect\citename{Kaplan and Kay}1994]{Kaplan+Kay:regmod}
Ronald~M. Kaplan and Martin Kay.
\newblock 1994.
\newblock Regular Models of Phonological Rule Systems.
\newblock {\em Computational Linguistics},  20:3, pages 331--378.

\bibitem[\protect\citename{Karttunen et al}1987]{KKK:twolcomp}
Lauri Karttunen, Kimmo Koskenniemi, and Ronald M. Kaplan.
\newblock 1987.
\newblock A Compiler for Two-level Phonological Rules.
\newblock In {\em Report No. CSLI-87-108. Center for the Study of
Language and Information}, Stanford University. Palo Alto, California.

\bibitem[\protect\citename{Karttunen}1994]{Karttunen:lt}
Lauri Karttunen.
\newblock 1994.
\newblock Constructing Lexical Transducers.
\newblock In {\em The Proceedings of the Fifteenth International
Conference on Computational Linguistics. Coling 94}, I, pages 406--411,
Kyoto, Japan.

\bibitem[\protect\citename{Karttunen}1995]{Karttunen:repl}
Lauri Karttunen.
\newblock 1995.
\newblock The Replace Operator.
\newblock In {\em The Proceedings of the 33rd Annual Meeting of the
Association for Computational Linguistics. ACL-95}, pages 16--23,
Boston, Massachusetts.

\bibitem[\protect\citename{Kempe and Karttunen}1996]{Kempe+Karttunen:tech}
Andr\'{e} Kempe and Lauri Karttunen.
\newblock 1996.
\newblock Parallel Replacement in the Finite-State Calculus.
\newblock In {\em The Proceedings of the Sixteenth International
Conference on Computational Linguistics. Coling 96}. Copenhagen,
Denmark.

\bibitem[\protect\citename{Leech}1995]{Leech:bnc}
Geoffrey Leech.
\newblock 1995.
\newblock {\em User's Guide to the British National Corpus}.
\newblock Lancaster University.

\bibitem[\protect\citename{Mohri}1994]{Mohri:fsa+nlp}
Mehryar Mohri.
\newblock 1994.
\newblock {\em On Some Applications of Finite-State Automata Theory to
Natural Language Processing}.
\newblock Technical Report 94-22. L'Institute Gaspard Monge.
Universit\'{e} de Marne-la-Vall\'{e}e. Noisy Le Grand.

\bibitem[\protect\citename{Roche}1993]{Roche:diss}
Emmanuel Roche.
\newblock 1993.
\newblock {\em Analyse syntaxique transformationelle du fran\c{c}ais par
transducteurs et lexique-grammaire}.
\newblock Doctoral dissertation, Universit\'{e} Paris 7.

\bibitem[\protect\citename{Roche and Schabes}1995]{Roche+Schabes:pos}
Emmanuel Roche and Yves Schabes.
\newblock 1995.
\newblock Deterministic Part-of-Speech Tagging.
\newblock {\em Computational Linguistics}, 21:2, pages 227--53.

\bibitem[\protect\citename{Schuetzenberger}1977]{Schuetzenberger:seq}
Marcel Paul Sch\"{u}tzenberger.
\newblock 1977.
\newblock Sur une variante des fonctions sequentielles.
\newblock {\em Theoretical Computer Science}, 4, pages 47--57.

\bibitem[\protect\citename{Silberztein}1993]{Silberztein:intex}
Max Silberztein.
\newblock 1993.
\newblock {\em Dictionnaires Electroniques et Analyse Lexicale du
Fran\c{c}ais---Le Syst\`{e}me INTEX}, Masson, Paris, France.

\end{thebibliography}
\end{document}